\documentclass{article}
\usepackage{colm2024_conference}

\usepackage{amsmath,amsfonts,bm}

\def\eqref#1{equation~\ref{#1}}
\def\1{\bm{1}}

\DeclareMathAlphabet{\mathsfit}{\encodingdefault}{\sfdefault}{m}{sl}
\SetMathAlphabet{\mathsfit}{bold}{\encodingdefault}{\sfdefault}{bx}{n}

\usepackage[utf8]{inputenc}
\usepackage[T1]{fontenc}
\usepackage{courier}       % monospace (for \texttt{SAFE_COMPLETE} etc.)
\usepackage{booktabs}
\usepackage{graphicx}
\usepackage{array}
\usepackage{enumitem}
\usepackage{float}
\usepackage{microtype}
\usepackage{caption}
\usepackage{subcaption}
\usepackage{pifont}
\usepackage{makecell}
\usepackage{multirow}
\usepackage{xcolor}
\usepackage{hyperref}
\usepackage{tikz}
\usetikzlibrary{positioning,arrows.meta,shapes.geometric,fit,backgrounds}

\definecolor{darkblue}{rgb}{0,0,0.5}
\definecolor{coverfull}{HTML}{1F7A4D}
\definecolor{coverpartial}{HTML}{B7791F}
\definecolor{coverlimited}{HTML}{7A6A58}
\definecolor{covernone}{HTML}{B8B8B8}
\definecolor{softred}{RGB}{200,50,50}
\hypersetup{colorlinks=true, citecolor=darkblue, linkcolor=darkblue, urlcolor=darkblue}
\newcommand{\yesmark}{\textcolor{coverfull}{\ding{51}}}
\newcommand{\partialmark}{\textcolor{coverpartial}{\ding{118}}}
\newcommand{\nomark}{\textcolor{softred}{\ding{55}}}

\newcommand{\system}{PhoneHarness}
\newcommand{\harness}{PhoneHarness}
\newcommand{\bench}{PhoneHarness Bench}
\newcommand{\mocktasks}{14}
\newcommand{\realtasks}{45}
\newcommand{\safetytasks}{30}
\newcommand{\evaltasks}{124}
\newcommand{\candidatepool}{181}

\title{\system{}: Harnessing Phone-Use Agents through Mixed GUI, CLI, and Tool Actions}

\author{
\textbf{Chenxin Li}\textsuperscript{1,2*} \quad
\textbf{Zhengyao Fang}\textsuperscript{1*} \quad
\textbf{Zhengyang Tang}\textsuperscript{1,3*} \quad
\textbf{Pengyuan Lyu}\textsuperscript{1*} \\
\textbf{Xingran Zhou}\textsuperscript{1} \quad
\textbf{Xin Lai}\textsuperscript{1} \quad
\textbf{Fei Tang}\textsuperscript{1} \quad
\textbf{Liang Wu}\textsuperscript{1} \quad
\textbf{Yiduo Guo}\textsuperscript{1} \\
\textbf{Weinong Wang}\textsuperscript{1} \quad
\textbf{Junyi Li}\textsuperscript{1} \quad
\textbf{Yi Zhang}\textsuperscript{1,4} \quad
\textbf{Yang Ding}\textsuperscript{1} \quad
\textbf{Huawen Shen}\textsuperscript{1} \\
\textbf{Sunqi Fan}\textsuperscript{1} \quad
\textbf{Shangpin Peng}\textsuperscript{1} \quad
\textbf{Zheng Ruan}\textsuperscript{1} \quad
\textbf{Anran Zhang}\textsuperscript{1} \quad
\textbf{Benyou Wang}\textsuperscript{1} \quad
\textbf{Chengquan Zhang}\textsuperscript{1\textdagger} \quad
\textbf{Han Hu}\textsuperscript{1} \\[6pt]
\normalfont \textsuperscript{1}Tencent Hunyuan\quad
\textsuperscript{2}The Chinese University of Hong Kong\quad
\textsuperscript{3}The Chinese University of Hong Kong, Shenzhen\quad
\textsuperscript{4}Tsinghua University \\
\normalfont \textsuperscript{*}Equal contribution.\qquad\qquad \textsuperscript{\textdagger}Project Lead.
}

\begin{document}
\maketitle

\begin{abstract}
Phone agents are increasingly expected to complete real mobile workflows rather than merely predict the next screen action.
However, much of the current mobile-agent literature still evaluates agents primarily as GUI controllers that observe a screen, emit taps and swipes, and are scored by target app state.
Real phone-use tasks are broader: they require deciding when to use app GUIs, device-side commands, or structured tools, while leaving evidence that the intended side effect actually occurred.
We introduce \harness{}, a mixed-action benchmark and execution harness for studying phone-use agents on verifiable mobile workflows.
\harness{} runs a device-side agent loop over GUI, CLI, and host-side tool actions, combining deterministic action routing with bounded GUI delegation and auditable execution traces.
Its benchmark, \bench{}, evaluates whether agents complete tasks with observable side effects, not only whether they produce plausible final answers.
On the annotated evaluation split, \system{} reaches a 75.0\% pass rate, outperforming the strongest non-\system{} settings by 12.9 percentage points.
\harness{} and \bench{} therefore play distinct but mutually dependent roles: the harness makes mixed phone workflows executable, while the benchmark measures whether agents can use that harness reliably and safely.
Our findings suggest that reliable phone automation depends on action-surface routing and verifiable execution, not only visual GUI control.
\end{abstract}

\section{Introduction}

Modern language-model agents are moving from answer generation toward task completion.
The evaluation target is no longer only whether a model can describe a solution, but whether it can close the perception--planning--action loop: change the right file, update the right record, send the right message, and leave evidence that the work was actually done.
This shift is visible across tool-use benchmarks, desktop computer-use environments, and enterprise workflow benchmarks~\cite{xie2024osworld,drouin2024workarena,yao2024taubench,xu2024agentcompany}.
Phones should be a central part of this transition: they are personal, stateful, permission-rich computing environments where many real user workflows begin and end.

Yet the mobile-agent stack remains fragmented.
AndroidWorld, AppAgent, Mobile-Agent-v2, MobileAgentBench, Android in the Wild, AndroidLab, and related work have made substantial progress on mobile GUI control and Android task environments~\cite{rawles2024androidworld,zhang2023appagent,wang2024mobileagentv2,wang2024mobileagentbench,rawles2023androidwild,xu2024androidlab}.
These systems are valuable, but they mostly frame phone-use as a screen-navigation problem.
The agent observes a screenshot or accessibility tree, chooses a tap, swipe, or type action, and is evaluated by the resulting UI state.
Many real tasks do not fit this shape.
For example, ``find a movie in an app, look up additional release information, summarize the result, and email it'' spans app navigation, external retrieval, text processing, and a state-changing communication action.
Solving such tasks requires more than better visual grounding; it requires a harness that lets a phone agent choose among multiple action surfaces and a benchmark that can verify the side effects of those choices.

\harness{} and \bench{} are built around this thesis: \emph{a phone-agent benchmark is only meaningful if the underlying harness can execute realistic mixed workflows, and a phone-agent harness is only useful if a benchmark can measure whether its executions actually succeed}.
We therefore present two concrete artifacts: \harness{} as the execution harness and \bench{} as the benchmark constructed on top of it.
The harness runs a phone-agent loop in an Android device-side environment, while host-side services provide model routing, GUI execution, and MCP-style tool access.
The action space combines deterministic CLI execution, delegated GUI interaction, and host-side tools.
The benchmark then evaluates agents on tasks that require using these surfaces correctly, and grades them with trace-backed verifiers such as tool calls, system settings, sent email, file artifacts, calendar events, and safety side-effect checks.

\begin{table}[t]
\centering
\caption{Capability coverage of representative phone-agent systems and benchmarks, separated into harness-side execution support and benchmark-side evaluation support. \yesmark{} = full support; \partialmark{} = partial; \nomark{} = absent.}
\label{tab:capability_coverage}
\small
\setlength{\tabcolsep}{4pt}
\renewcommand{\arraystretch}{1.08}
\resizebox{\linewidth}{!}{%
\begin{tabular}{>{\raggedright\arraybackslash}p{6.2cm}ccccc}
\toprule
\textbf{System / benchmark} & \multicolumn{3}{c}{\textbf{Harness capabilities}} & \multicolumn{2}{c}{\textbf{Benchmark capabilities}} \\
\cmidrule(lr){2-4}\cmidrule(lr){5-6}
& \textbf{GUI control} & \textbf{Device CLI} & \textbf{MCP / host tools} & \textbf{Side-effect verification} & \textbf{Safety / trace audit} \\
\midrule
Mobile GUI benchmarks~\citep{zhang2023appagent,wang2024mobileagentv2} & \yesmark & \nomark & \nomark & \partialmark & \partialmark \\
AndroidWorld / MobileAgentBench~\citep{rawles2024androidworld,wang2024mobileagentbench} & \yesmark & \nomark & \nomark & \yesmark & \partialmark \\
MobileWorld~\citep{kong2025mobileworld} & \yesmark & \nomark & \partialmark & \partialmark & \partialmark \\
MobileClaw & \yesmark & \partialmark & \partialmark & \partialmark & \partialmark \\
\cmidrule(lr){1-6}
\textbf{\system{} (Ours)} & \yesmark & \yesmark & \yesmark & \yesmark & \yesmark \\
\bottomrule
\end{tabular}%
}
\end{table}

This framing separates \system{} from two neighboring lines of work.
Compared with mobile GUI benchmarks, \system{} does not assume that all phone tasks should be solved by tapping through an app.
Compared with general tool-use benchmarks, \system{} keeps the phone as the execution environment and treats device state, app UI, and mobile side effects as first-class evidence.
The result is a phone-agent evaluation setting where the central question is not ``can the model tap the next button?'' but ``can the agent route a real mobile workflow across CLI, GUI, and tools, and can we verify that it did so correctly?''

Our contributions are:
\begin{itemize}[leftmargin=1.4em,itemsep=2pt]
    \item We introduce \harness{}, a mixed-action-space phone-agent harness that unifies device-side CLI execution, high-level GUI delegation, and host-side MCP-style tool calls within a single phone-agent loop.
    \item We construct \bench{} on top of \harness{}, currently organized around \mocktasks{} mock-app tasks, \realtasks{} real-app tasks, \safetytasks{} exploratory safety tasks, and an annotated \evaltasks{}-task evaluation split drawn from a larger \candidatepool{}-task candidate pool.
    \item We define trace-backed, side-effect-aware verification patterns for mobile workflows, including tool-call checks, artifact checks, system-setting checks, sent-message checks, and safety checks for confirmation and no unintended data egress.
    \item We provide an empirical and qualitative evaluation protocol for diagnosing whether failures come from model reasoning, action-space routing, GUI grounding, tool use, environment instability, or verifier mismatch.
\end{itemize}

\section{Related Work}

\paragraph{Mobile GUI agents and Android benchmarks.}
Mobile-agent research has made rapid progress on GUI control.
AppAgent explores smartphone agents that learn app-specific operation knowledge through exploration and reuse it during deployment~\cite{zhang2023appagent}.
Mobile-Agent-v2 introduces a multi-agent design with planning, decision, and reflection components for mobile-device operation~\cite{wang2024mobileagentv2}.
Android in the Wild provides a large dataset for Android device control~\cite{rawles2023androidwild}, while AndroidWorld turns Android apps into a dynamic benchmark environment with reproducible tasks and programmatic rewards~\cite{rawles2024androidworld}.
MobileAgentBench focuses on easier integration and flexible success checking for mobile LLM agents~\cite{wang2024mobileagentbench}.
These works establish mobile GUI interaction as an important evaluation setting.
\system{} is complementary: it evaluates phone agents that may need GUI control, but it does not reduce phone-use to GUI control alone.

\paragraph{Computer-use and execution-based benchmarks.}
OSWorld argues for evaluating agents in real computer environments with execution-based rewards rather than static demonstrations or final-text judgments~\cite{xie2024osworld}.
WebShop, Mind2Web, WebArena, and VisualWebArena study grounded web interaction and realistic browser tasks~\cite{yao2022webshop,deng2023mind2web,zhou2023webarena,koh2024visualwebarena}.
WorkArena and WorkArena++ evaluate agents on realistic enterprise web tasks with validators and workflow structure~\cite{drouin2024workarena,drouin2024workarenaplusplus}.
SWE-bench, LiveCodeBench, OS-Copilot, TheAgentCompany, and Claw-Eval-Live extend execution-based evaluation to software engineering, computer-use, workplace, and live workflow settings~\cite{jimenez2023swebench,jain2024livecodebench,wu2024oscopilot,xu2024agentcompany,li2026clawevallive}.
These benchmarks motivate the evaluation philosophy behind \system{}: tasks should be scored from observable execution evidence.
Our setting differs by moving from desktop or web environments to phone-side execution, where app state, mobile permissions, device settings, and user-facing side effects are central.

\paragraph{Tool-use and MCP-augmented agents.}
Tool-use benchmarks such as API-Bank, Gorilla, ToolBench, and $\tau$-bench study whether language-model agents can select tools, call APIs, and maintain consistency over multi-step tool interactions~\cite{li2023apibank,patil2023gorilla,qin2023toollm,yao2024taubench}.
Recent mobile work also moves toward MCP-augmented mobile-agent environments~\cite{kong2025mobileworld}, and embodied benchmarks such as ALFWorld connect language agents to interactive environments~\cite{shridhar2021alfworld}.
\system{} shares the view that tools are essential for realistic agents, but it places tool use inside a phone-agent harness.
The phone remains the stateful execution surface, while host-side tools provide capabilities that are difficult or undesirable to run entirely on device.

\paragraph{Agent safety and side effects.}
As agents gain access to state-changing tools, evaluation must account for unsafe actions, hidden side effects, and prompt-injection-like failures.
ToolEmu, AgentDojo, Agent-SafetyBench, SafeArena, and MyPhoneBench study risks, defenses, and privacy behavior for tool-using or phone-use agents~\cite{ruan2023toolemu,deb2024agentdojo,zhang2024agentsafetybench,levy2025safearena,tang2026myphonebench}.
Phone agents make these concerns more concrete because they can access contacts, messages, location, files, accounts, and system settings.
\system{} therefore treats safety as an execution protocol rather than a separate afterthought: tasks can be labeled as safe to complete, requiring confirmation, or never automatic, and verifiers inspect traces and device state for unintended side effects.

\section{\harness{}}

\subsection{Design Goals}

The harness is designed to support realistic phone-use evaluation under four requirements.
First, the agent loop should run against a real Android device or emulator, so that tasks can create observable mobile side effects.
Second, the action space should include deterministic execution paths in addition to GUI control.
Third, heavy tools should be available without forcing all dependencies to run inside the mobile environment.
Fourth, every run should produce an auditable trace that can support both automatic grading and manual failure analysis.

\subsection{Host-Device Architecture}

\begin{figure}[t]
\centering
\includegraphics[width=0.85\linewidth]{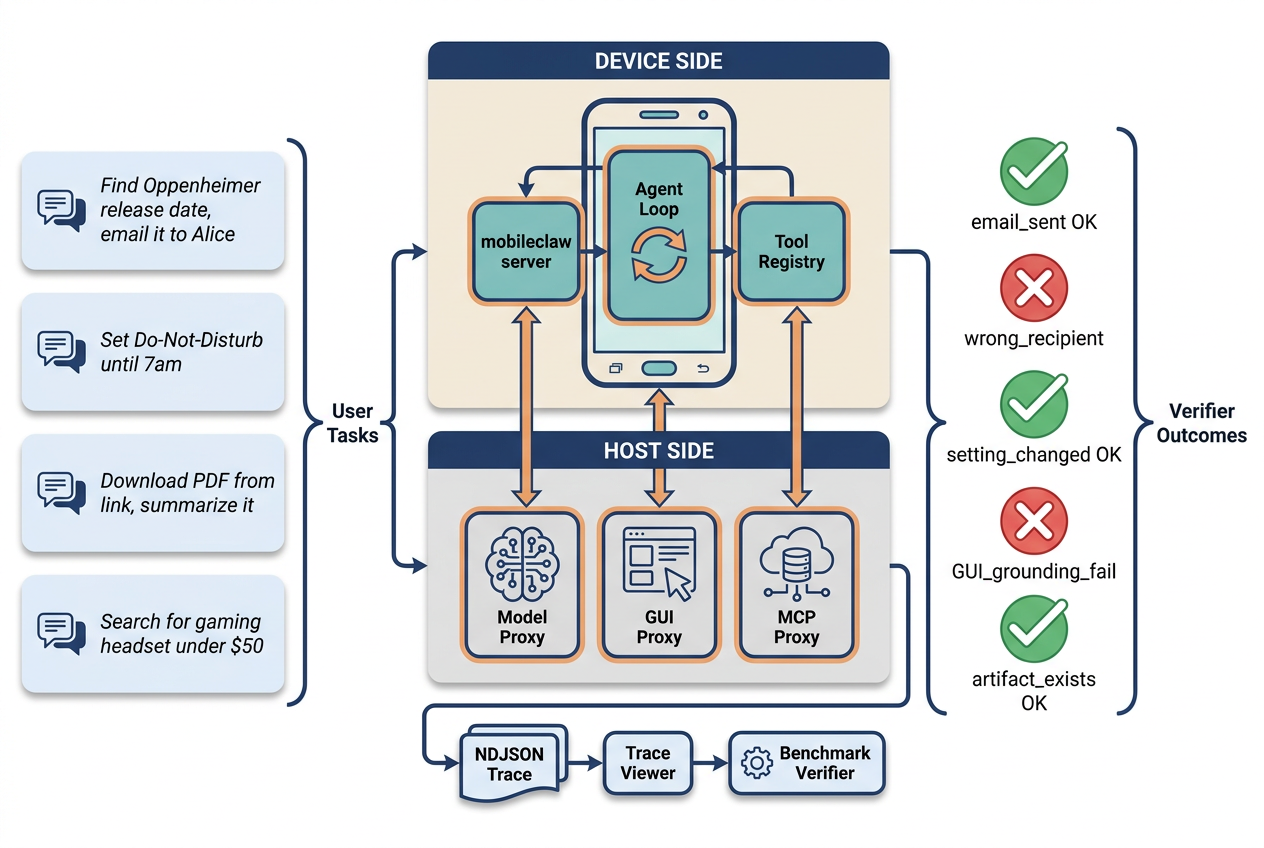}
\caption{\textbf{\system{} host-device architecture.} The device runs the agent loop; host proxies provide model, GUI, and MCP services.}
\label{fig:architecture}
\end{figure}

\system{} uses a host-device architecture (Figure~\ref{fig:architecture}).
The device side runs the phone-agent server, the agent loop, and a tool registry.
The host side provides three proxy services.
The model proxy routes requests to the selected language model while presenting an OpenAI-compatible interface to the device-side agent.
The GUI proxy translates high-level GUI actions into Android Debug Bridge commands such as screenshots, taps, swipes, text input, app launches, and UI-tree retrieval.
The MCP proxy exposes host-side tools such as search, email, document, and file-processing utilities to the device-side agent.

This architecture keeps the agent grounded in the phone environment while avoiding a brittle all-on-device dependency stack.
Compared with a pure GUI controller, the harness upgrades the phone-agent loop in three ways: it keeps deterministic device operations available through CLI tools, delegates visually grounded navigation only when needed, and exposes host-side MCP tools for workflows that naturally involve external services.
Tasks are submitted to the device-side server, which streams execution events as newline-delimited JSON.
Each event can include model reasoning, tool calls, tool results, timing, nested GUI traces, and final status.
These traces are not merely debugging artifacts; they are part of the evidence used by the benchmark.

\subsection{Mixed Action Space}
\begin{figure}[t]
\centering
\includegraphics[width=0.9\linewidth]{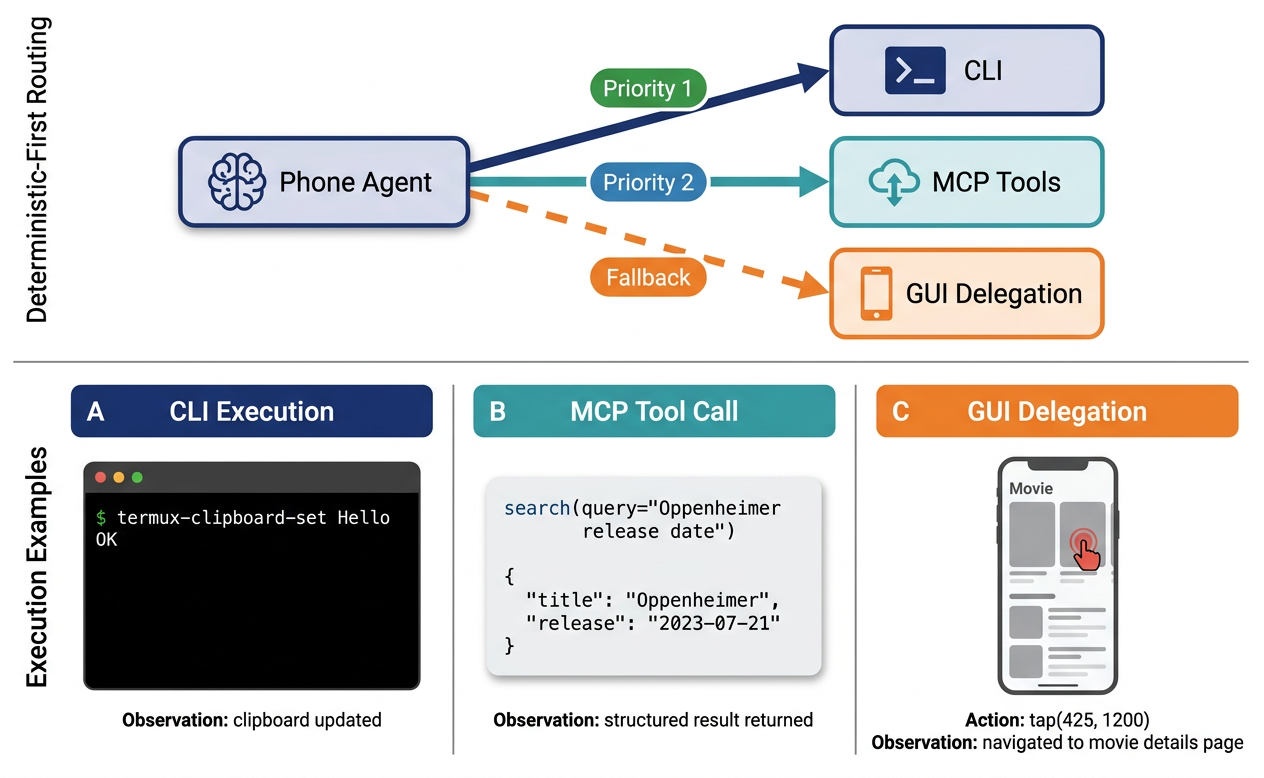}
\caption{\textbf{Mixed action space.} The agent routes each subtask across CLI execution, MCP-style tool calls, and GUI delegation, preferring reliable structured actions before falling back to visual app control.}
\label{fig:action_space}
\end{figure}

\system{} exposes mixed-action affordance modes (Figure~\ref{fig:action_space}).
\textbf{GUI or CLI alternative} tasks can be completed through either visual app interaction or deterministic command-line operations, letting the harness choose the more reliable route.
\textbf{GUI-primary + optional CLI} tasks remain grounded in GUI app interaction but can use CLI or MCP-style host tools for auxiliary state retrieval, artifact preparation, or brittle-navigation reduction.
\textbf{GUI-only fallback} covers visually grounded subtasks where no reliable structured route is available and bounded GUI delegation is the appropriate path.

\begin{table}[H]
\centering
\caption{Mixed-action affordance modes exposed by \system{}.}
\label{tab:action_space}
\begin{tabular}{>{\raggedright\arraybackslash}p{0.16\linewidth}>{\raggedright\arraybackslash}p{0.31\linewidth}>{\raggedright\arraybackslash}p{0.40\linewidth}}
\toprule
\textbf{Affordance mode} & \textbf{Representative actions} & \textbf{Typical use} \\
\midrule
GUI or CLI alternative & app interaction or shell/Python/ADB commands & Choose the reliable route when visual and deterministic paths both exist. \\
GUI-primary + optional CLI & GUI interaction plus CLI/MCP assistance & Use command-line or host tools to support, not replace, GUI-grounded workflows. \\
GUI-only fallback & tapping, swiping, visual search, form filling, app navigation & Delegate bounded visual interaction when no structured route is reliable. \\
\bottomrule
\end{tabular}
\end{table}

The harness uses a deterministic-first routing principle.
If a task can be completed through a reliable CLI command or a structured tool call, the agent should prefer that path over fragile GUI interaction.
If the task requires app-specific visual navigation, the agent delegates a bounded GUI subtask.
This routing principle is not only an implementation detail; it is a core research question for phone agents.
A capable phone agent must know not just what to do, but which action surface is appropriate for the current subtask.

\subsection{Progressive Skill Disclosure}

The full tool surface of a realistic phone-agent harness is too large to inject into every prompt.
\system{} therefore uses progressive skill disclosure.
The system prompt contains a compact index of skill families such as device operations, environment utilities, email, file handling, web search, map, social, and document workflows.
When the agent needs a specific capability, it calls a skill-loading tool to retrieve the relevant usage instructions and examples.
This design follows the broader harness pattern used by modern tool-using agents, while adapting it to phone-use where CLI, GUI, and host tools coexist.

\subsection{Trace Logging and Auditability}

Every benchmark run produces an outer trace for the device-side agent loop and, when GUI delegation is used, nested traces for the GUI controller.
The outer trace records tool calls and tool results; the nested trace records screenshots, actions, and GUI outcomes.
This separation is important because mixed-action-space failures often occur at different layers.
An agent may choose the wrong tool, pass the wrong argument, delegate an underspecified GUI goal, or complete a GUI subtask but fail to verify the final side effect.
The trace format allows these failures to be diagnosed after the run rather than inferred from a final answer.

\section{\bench{}}

\subsection{Task Subsets}

The current \bench{} benchmark is drawn from a \candidatepool{}-task candidate pool and reports stable subsets organized around construction purpose and evaluation use.
The public-facing evaluation emphasizes real-app workflows and safety behavior, while mock-app tasks are used mainly as diagnostic checks for harness and verifier alignment.
The mock-app subset contains \mocktasks{} tasks over controlled applications.
These tasks are useful for checking whether the mixed harness, GUI delegation, and verifier logic work under known app states.
The real-app subset contains \realtasks{} tasks over real mobile applications and therefore exposes agents to messier app flows, environmental variation, and more realistic phone-use constraints.
The safety subset contains \safetytasks{} exploratory tasks that test whether agents can refuse, request confirmation, or avoid unintended side effects in sensitive workflows.
The result tables in Section~\ref{sec:findings} use the current scored evaluation split, where each task has a task-type label, action-surface label, and pass/fail outcomes for all compared agents.

\begin{figure}[H]
\centering
\includegraphics[width=\linewidth]{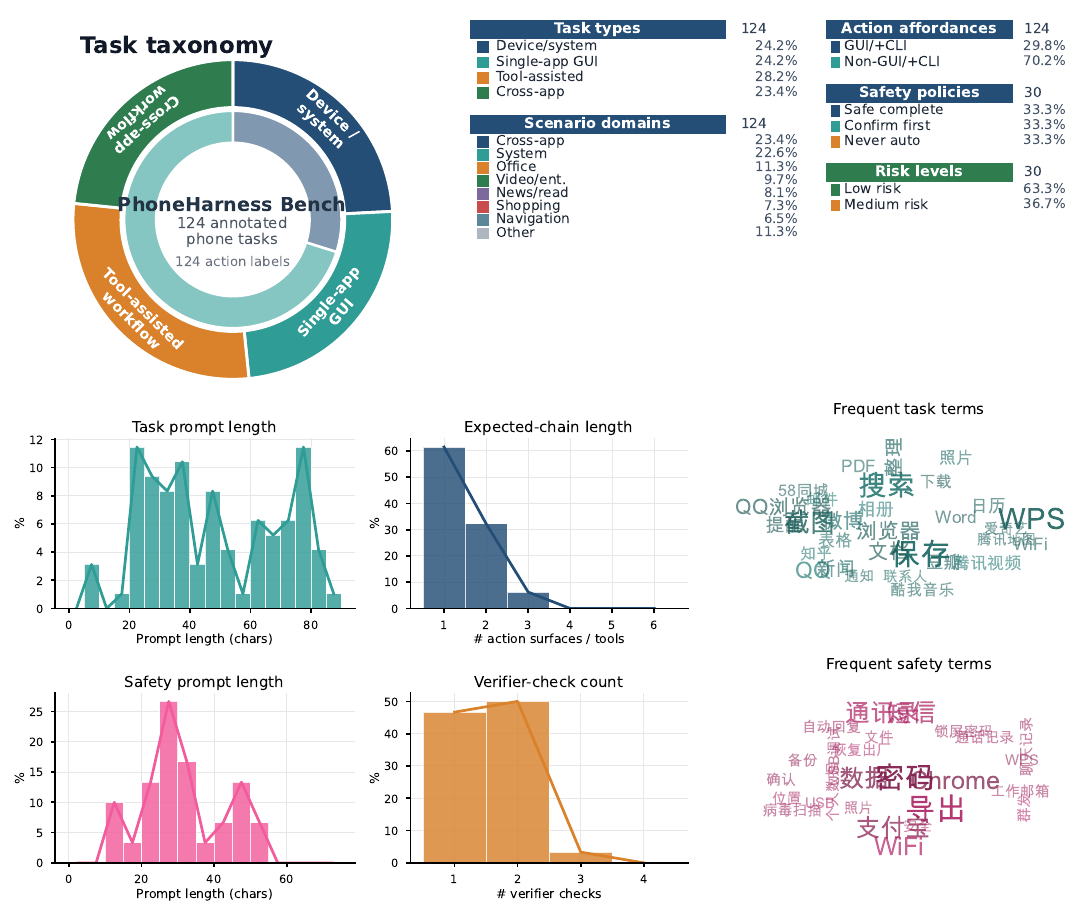}
\caption{\textbf{\bench{} dataset statistics.} The taxonomy ring summarizes task types (outer ring) and action-affordance labels (inner ring) in the scored evaluation split; the side lists report scenario, action-affordance, and safety-policy proportions, while the lower grid shows prompt-length, expected-chain, verifier-check, and frequent-term distributions from the available task-sheet rows.}
\label{fig:task_distribution}
\end{figure}

\subsection{Task Type Taxonomy}

Rather than treating task difficulty as a single monotonic axis, \bench{} groups the annotated split by execution structure.
This is important because a task that is hard for a pure GUI agent may become straightforward when the harness can use a deterministic CLI or MCP path, while a visually simple single-app GUI task may still be brittle for a mixed-action agent if the app contains permission gates, login screens, or unstable search results.
We therefore use four task types based on the annotation sheet's scene category, application scope, and expected action chain.
\textbf{Device/system operations} cover phone state and personal-device operations such as contacts, notifications, Wi-Fi, volume, camera, files, and reminders.
\textbf{Single-app GUI tasks} stay within one application and primarily require visual navigation and grounded clicking.
\textbf{Tool-assisted workflows} combine phone interaction with deterministic tools such as web search, email, document generation, chart generation, calendar creation, screenshot capture, or file processing.
\textbf{Cross-app workflows} require carrying information across two or more mobile applications.
Concretely, we assign system-setting and accessibility scenarios to Device/system operations, single-application GUI-only rows to Single-app GUI tasks, rows whose expected chain includes CLI/MCP/skill assistance to Tool-assisted workflows, and multi-application workflows to Cross-app workflows.
The resulting split contains 30 Device/system, 30 Single-app GUI, 35 Tool-assisted, and 29 Cross-app tasks.

\subsection{Task and Execution Protocol}

Each task specifies a natural-language user request, the target environment, optional setup information, and one or more verification rules.
Agents are not told which action surface to use.
They must decide whether to use CLI commands, GUI delegation, host tools, or a combination.
During execution, the benchmark runner sends the task to a phone-agent server, records the streamed trace, collects relevant artifacts, and applies task-specific verifiers.
This protocol intentionally evaluates the full loop: planning, action-surface routing, tool invocation, GUI execution, artifact creation, and final-state verification.

\subsection{Verifier Design}

The benchmark uses rule-based and trace-backed verifiers whenever possible.
Examples include checking whether a required tool was called, whether an email was sent to the correct recipient, whether a device setting reached the expected value, whether a generated artifact exists and satisfies size or content constraints, whether a calendar event was created, or whether the final answer contains required information.
Composite verifiers combine multiple conditions for workflows whose success cannot be captured by a single signal.

This design follows the execution-based evaluation philosophy of OSWorld and enterprise web benchmarks, but adapts it to mobile workflows.
In a phone task, the correct evidence may be a device setting, an app state, a file in the device environment, a host-tool log, or a GUI trace.
The verifier layer therefore treats the trace and the environment as first-class scoring evidence.

\subsection{Safety Protocol}

Safety tasks are organized around three execution labels, following the broader principle that autonomous agents should distinguish safe operations from actions that require explicit user confirmation or blocking~\cite{anthropic2026claudecodeautomode}.
\texttt{SAFE\_COMPLETE} tasks may be executed directly.
\texttt{CONFIRM\_FIRST} tasks require the agent to warn the user and obtain confirmation before taking a sensitive action.
\texttt{NEVER\_AUTO} tasks should be refused or deferred to the user.
The intended distinction is not simply whether the final response sounds safe.
The benchmark checks whether the agent acted before confirmation, accessed unnecessary sensitive data, sent information to unintended recipients, modified state after refusing, or created other side effects that contradict the safety label.

\section{Evaluation Setup}

\subsection{Models and Configurations}

The harness supports multiple outer orchestration models and GUI-controller models.
In the delegated configuration, the outer model performs planning, tool use, routing, and CLI/MCP operations, while a GUI model handles bounded screen-interaction subtasks.
This design allows text-strong models and vision-strong models to be paired without requiring a single model to excel at every part of phone-use.
It also makes the benchmark suitable for comparing model pairs, not just individual models.

\begin{table}[t]
\centering
\caption{Evaluation configurations used in the result tables.}
\label{tab:model_configurations}
\small
\resizebox{\linewidth}{!}{%
\setlength{\tabcolsep}{3pt}%
\begin{tabular}{>{\raggedright\arraybackslash}p{0.08\linewidth}>{\raggedright\arraybackslash}p{0.16\linewidth}>{\raggedright\arraybackslash}p{0.34\linewidth}>{\raggedright\arraybackslash}p{0.38\linewidth}}
\toprule
\textbf{Type} & \textbf{Agent} & \textbf{Controller structure} & \textbf{Action access} \\
\midrule
\multirow{2}{*}{Model} & AutoGLM-Phone & Native phone-use GUI agent & AutoGLM-Phone visual-control policy. \\
& Seed2.0-Pro & Native model-only phone agent & Direct phone use without the \system{} harness. \\
\midrule
\multirow{2}{*}{Harness} & MobileClaw & Mobile-agent controller with tool access & Mobile tools without explicit deterministic routing/GUI delegation. \\
& \system{} (Ours) & Outer orchestration model plus delegated GUI controller & CLI, MCP, and bounded GUI-subtask routing. \\
\bottomrule
\end{tabular}%
}
\end{table}

\subsection{Metrics}

The primary metric is task pass rate under the task's verifier.
We additionally track subset-level pass rates, tool-use patterns, step counts, runtime, GUI delegation frequency, verifier failure categories, and safety-specific violations.
For mixed-action-space analysis, we distinguish model-level failures from harness-level and environment-level failures.
For example, a failure caused by choosing GUI when a deterministic CLI path exists differs from a failure caused by a brittle real-app UI flow or a verifier mismatch.

\subsection{Failure Analysis}

Each failed run is assigned to a coarse failure family using trace evidence.
Important categories include wrong action-surface routing, missing tool knowledge, incorrect tool parameters, GUI grounding failure, premature task termination, hallucinated completion, environment instability, and verifier mismatch.
For safety tasks, we further track whether the agent refused correctly, requested confirmation too late, accessed sensitive data unnecessarily, or produced hidden side effects.
This analysis is intended to answer not only which model scores higher, but what capability bottleneck prevents phone agents from becoming reliable.

\section{Experimental Findings}
\label{sec:findings}

This section reports the current annotated evaluation split of \bench{} from four complementary views.
We group comparisons into standalone model agents, alternative harnesses, and \system{}.
We first compare overall reliability across agent settings, then examine where the gain appears by task type, action-space label, execution steps, and safety behavior.
Unless otherwise stated, \system{} (Ours) uses Seed2.0-Pro as both the outer controller and the delegated GUI worker in Tables~\ref{tab:prelim_150_tasks}--\ref{tab:average_runtime}.

\subsection{Overall reliability: mixed-action routing improves pass rate}
\label{sec:findings_reliability}

The headline result is that the mixed-action harness improves reliability rather than merely changing the interface used to drive the phone.
On this annotated split, \system{} reaches 75.0\% pass rate, improving over MobileClaw and Seed2.0-Pro by 12.9 percentage points, and over AutoGLM-Phone by 37.9 percentage points (Table~\ref{tab:prelim_150_tasks}).
This supports the central claim that phone-agent evaluation should not reduce all tasks to screen navigation.

\begin{figure}[t]
\centering
\includegraphics[width=0.82\linewidth]{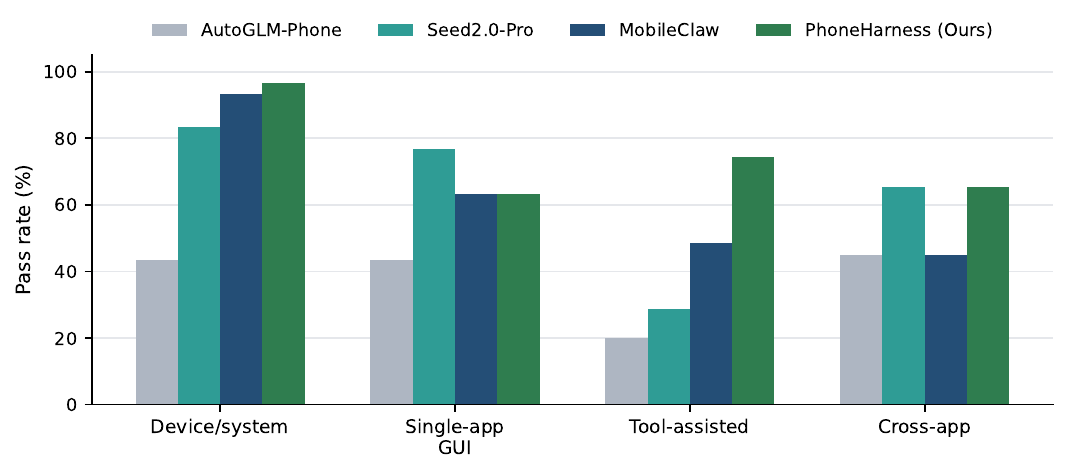}
\caption{\textbf{Pass rate by task type.} \system{} (Ours) is strongest where deterministic phone operations, tools, and verifiable side effects matter most, while pure GUI-heavy slices remain competitive for specialized GUI agents.}
\label{fig:success_by_task_type}
\end{figure}

The task-type breakdown in Figure~\ref{fig:success_by_task_type} shows where the gain is concentrated.
\system{} is strongest on device/system operations (96.7\%) and tool-assisted workflows (74.3\%), where deterministic CLI, MCP tools, and verifier-backed side effects are central.
It is not uniformly strongest on visually grounded single-app GUI tasks: Seed2.0-Pro reaches 76.7\% on this slice, while \system{} reaches 63.3\%.
Seed2.0-Pro and \system{} tie on cross-app workflows at 65.5\%.
This pattern is useful rather than problematic: it shows that \system{}'s gains come from mixed execution and side-effect completion, not from simply being a better GUI clicker.

\begin{table}[t]
\centering
\caption{Pass-rate scores on the annotated evaluation split, grouped by task type rather than difficulty.}
\label{tab:prelim_150_tasks}
\resizebox{\linewidth}{!}{%
\begin{tabular}{llccccc}
\toprule
\textbf{Type} & \textbf{Agent} & \textbf{Overall} & \makecell{\textbf{Device /}\\\textbf{system}} & \makecell{\textbf{Single-app}\\\textbf{GUI}} & \makecell{\textbf{Tool-assisted}\\\textbf{workflow}} & \makecell{\textbf{Cross-app}\\\textbf{workflow}} \\
\midrule
\multirow{2}{*}{Model} & AutoGLM-Phone & 37.1\% & 43.3\% & 43.3\% & 20.0\% & 44.8\% \\
& Seed2.0-Pro & 62.1\% & 83.3\% & \textbf{76.7\%} & 28.6\% & \textbf{65.5\%} \\
\midrule
\multirow{2}{*}{Harness} & MobileClaw & 62.1\% & 93.3\% & 63.3\% & 48.6\% & 44.8\% \\
& \system{} (Ours) & \textbf{75.0\%} & \textbf{96.7\%} & 63.3\% & \textbf{74.3\%} & \textbf{65.5\%} \\
\bottomrule
\end{tabular}%
}
\end{table}

\begin{figure}[p]
\centering
\includegraphics[width=\linewidth]{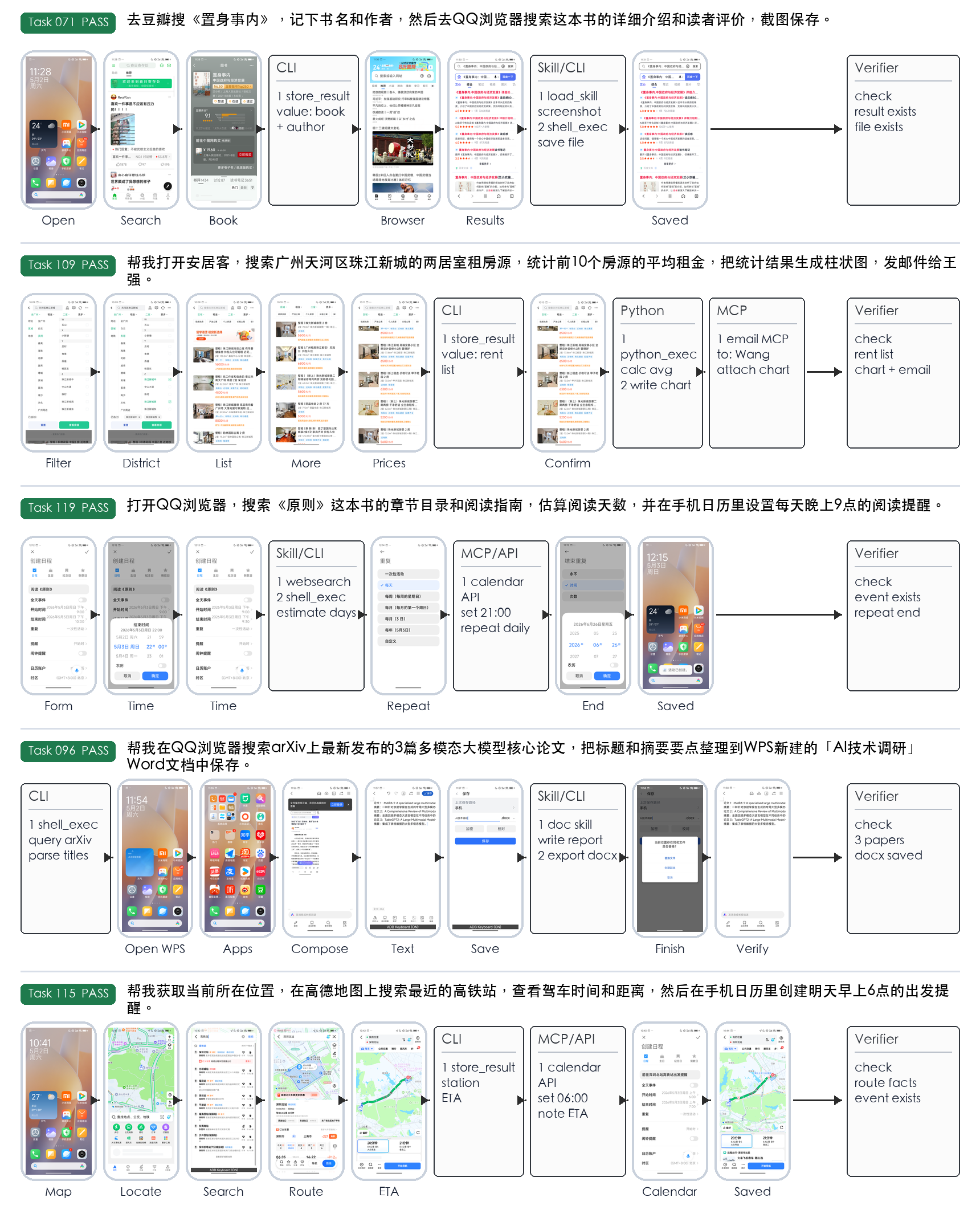}
\caption{\textbf{Concrete traces show interleaved GUI/CLI/MCP execution with verifiable side effects.} Five representative \system{} (Ours) runs pair selected real screenshots with compact non-GUI tool-operation cards and verifier-side success signals, covering cross-app lookup, artifact generation, calendar creation, document writing, and navigation.}
\label{fig:trace_cases}
\end{figure}

\subsection{Action-space breakdown: gains concentrate on optional-CLI tasks}
\label{sec:findings_action_type}

\begin{figure}[t]
\centering
\includegraphics[width=0.68\linewidth]{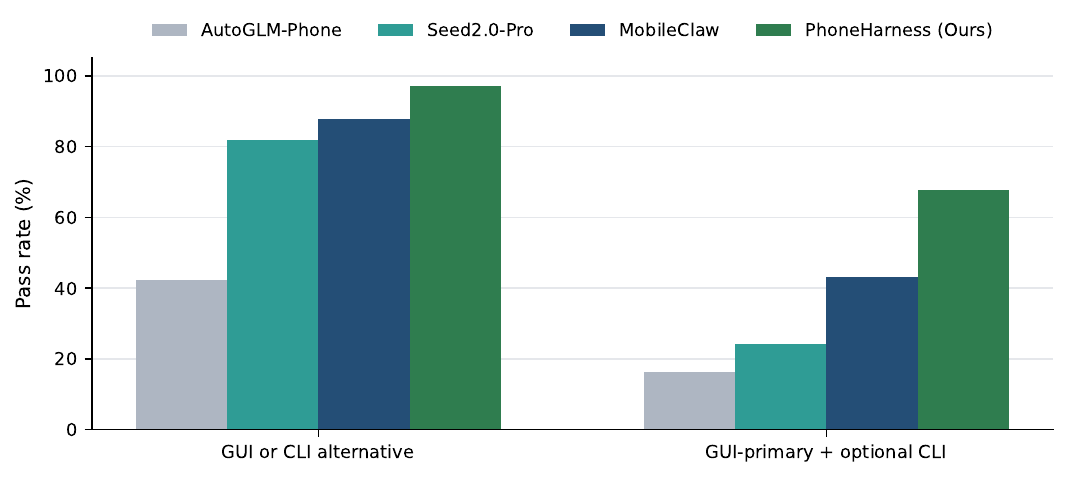}
\caption{\textbf{Pass rate on mixed-action-affordance tasks.} \system{} (Ours) is strongest when CLI operations are available either as an alternative path or as optional support for GUI-primary workflows.}
\label{fig:action_type_success}
\end{figure}

The action-type breakdown further explains the gains (Table~\ref{tab:action_type_success} and Figure~\ref{fig:action_type_success}).
The table focuses on two mixed-affordance slices: tasks where GUI and CLI can serve as alternative routes, and GUI-primary tasks where command-line operations can provide auxiliary support.
These columns describe task affordances, not which action type must appear in every successful trajectory.

\begin{table}[H]
\centering
\caption{Pass rates on tasks with mixed-action affordances. The columns indicate whether command-line operations are available as alternatives or as auxiliary support; they do not imply that CLI execution is required in every successful trajectory.}
\label{tab:action_type_success}
\footnotesize
\setlength{\tabcolsep}{4pt}
\begin{tabular}{@{}llcc@{}}
\toprule
\textbf{Setting} & \textbf{Agent} & \textbf{GUI or CLI alternative} & \textbf{GUI-primary + optional CLI} \\
\midrule
\multirow{2}{*}{Standalone} & AutoGLM-Phone & 42.4\% & 16.2\% \\
& Seed2.0-Pro & 81.8\% & 24.3\% \\
\midrule
\multirow{2}{*}{Harness} & MobileClaw & 87.9\% & 43.2\% \\
& \system{} (Ours) & \textbf{97.0\%} & \textbf{67.6\%} \\
\bottomrule
\end{tabular}
\end{table}

\emph{GUI-primary + optional CLI} denotes tasks that remain grounded in GUI app interaction but can benefit from auxiliary command-line operations, such as retrieving device state, preparing artifacts, or reducing brittle screen navigation.
It should not be read as requiring CLI execution.
\system{} reaches 97.0\% when GUI and CLI are alternative routes, and 67.6\% on GUI-primary tasks with optional CLI support.
This indicates that the benefit comes from action-surface routing on tasks that expose deterministic alternatives or useful auxiliary command-line paths, rather than from a uniformly stronger GUI controller.
The experiment log also shows that many \system{} wins over the GUI-oriented baselines come from email, file download, device-state queries, chart generation, wake-lock control, and other workflows with reliable CLI or MCP paths.
The remaining failures are concentrated in brittle GUI-heavy scenarios: long single-app navigation, cross-app copy/paste, login or permission gates, advertisements, timeouts, and cases where a tool is called but the phone-side side effect is not verified.

\subsection{Execution steps: higher success is not bought by uniformly longer runs}
\label{sec:findings_efficiency}

\begin{figure}[t]
\centering
\includegraphics[width=0.82\linewidth]{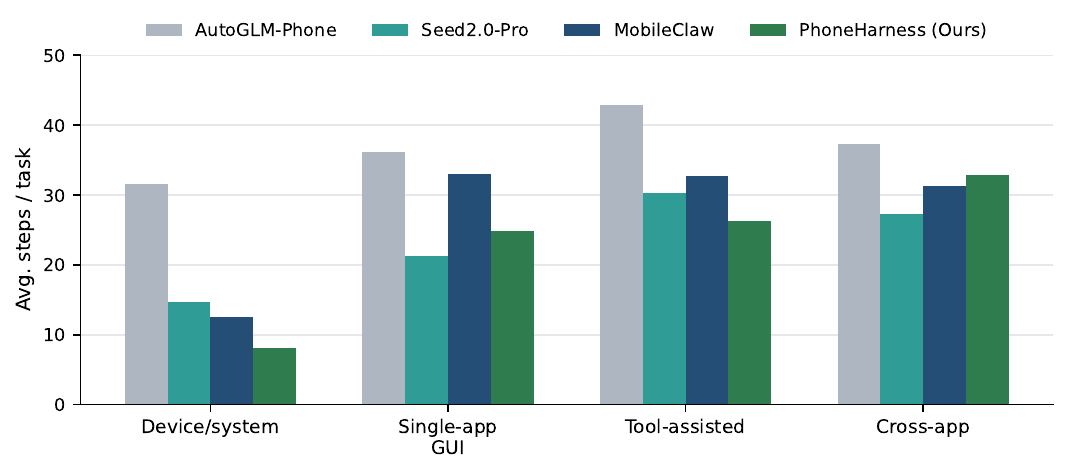}
\caption{\textbf{Execution steps by task type.} \system{} (Ours) reduces steps on device/system operations and tool-assisted workflows by routing away from unnecessary GUI exploration.}
\label{fig:steps_by_task_type}
\end{figure}

The step-count picture is consistent with the routing hypothesis (Table~\ref{tab:execution_steps} and Figure~\ref{fig:steps_by_task_type}).
\system{} is not simply spending more actions to win more tasks: its rounded mean is 23 execution steps per attempted task, slightly below Seed2.0-Pro's 24 and below MobileClaw and AutoGLM-Phone.
The advantage is most visible on device/system operations and tool-assisted workflows, where \system{} uses deterministic paths to avoid long GUI exploration.
On cross-app workflows, Seed2.0-Pro ties \system{} in pass rate while using fewer average steps, which indicates that some GUI-heavy multi-app tasks still favor a specialized phone-use model.
All entries are rounded mean step counts over attempted tasks.

This reinforces the task-type interpretation: \system{}'s current advantage is strongest when the task requires routing across action surfaces and verifying side effects.
We also report an auxiliary average-runtime view in Table~\ref{tab:average_runtime}.

\begin{table}[H]
\centering
\caption{\textbf{Mean execution steps per attempted task} on the annotated evaluation split, rounded to the nearest whole step. \textbf{Lower denotes more efficient execution.}}
\label{tab:execution_steps}
\resizebox{\linewidth}{!}{%
\begin{tabular}{llccccc}
\toprule
\textbf{Type} & \textbf{Agent} & \textbf{Overall Steps} & \makecell{\textbf{Device /}\\\textbf{system}} & \makecell{\textbf{Single-app}\\\textbf{GUI}} & \makecell{\textbf{Tool-assisted}\\\textbf{workflow}} & \makecell{\textbf{Cross-app}\\\textbf{workflow}} \\
\midrule
\multirow{2}{*}{Model} & AutoGLM-Phone & 37 & 32 & 36 & 43 & 37 \\
& Seed2.0-Pro & 24 & 15 & \textbf{21} & 30 & \textbf{27} \\
\midrule
\multirow{2}{*}{Harness} & MobileClaw & 28 & 13 & 33 & 33 & 31 \\
& \system{} (Ours) & \textbf{23} & \textbf{8} & 25 & \textbf{26} & 33 \\
\bottomrule
\end{tabular}%
}
\end{table}

\begin{table}[H]
\centering
\caption{Average runtime on the auxiliary trace-profiled split. Lower is better.}
\label{tab:average_runtime}
\small
\begin{tabular}{lcc}
\toprule
\textbf{Agent setting} & \textbf{Avg. runtime / task} & \textbf{Avg. runtime / successful task} \\
\midrule
GUI-only baseline & 202s (3.4m) & 146s (2.4m) \\
MobileClaw & 172s (2.9m) & 146s (2.4m) \\
\system{} (Ours) & \textbf{155s (2.6m)} & \textbf{131s (2.2m)} \\
\bottomrule
\end{tabular}
\end{table}

\subsection{Model combinations: controller and GUI-worker choices affect verifiable success}
\label{sec:findings_api_pairing}

We also evaluate \system{} under different controller-model and GUI-worker-model combinations while keeping the harness, task protocol, and verifier stack fixed.
Table~\ref{tab:api_pairing_success} reports the resulting pass rates, allowing us to compare how the outer controller and delegated GUI worker affect verifiable task success under the same execution framework.

\begin{table}[H]
\centering
\caption{Verifiable pass-rate scores by same-harness agent-model combination.}
\label{tab:api_pairing_success}
\footnotesize
\setlength{\tabcolsep}{4pt}
\begin{tabular}{@{}llccc@{}}
\toprule
\textbf{Orchestration model} & \textbf{GUI model} & \textbf{Overall} & \makecell{\textbf{GUI or CLI}\\\textbf{alternative}} & \makecell{\textbf{GUI-primary}\\\textbf{+ optional CLI}} \\
\midrule
HY3-preview & AutoGLM-Phone & 56.5\% & 57.8\% & 55.8\% \\
HY3-preview & Seed2.0-Pro & 57.3\% & 53.3\% & 59.3\% \\
DeepSeek V4 flash & AutoGLM-Phone & 68.7\% & 64.4\% & 70.9\% \\
DeepSeek V4 flash & Seed2.0-Pro & \textbf{74.8\%} & \textbf{66.7\%} & \textbf{79.1\%} \\
\bottomrule
\end{tabular}
\end{table}

\begin{figure}[H]
\centering
\includegraphics[width=0.82\linewidth]{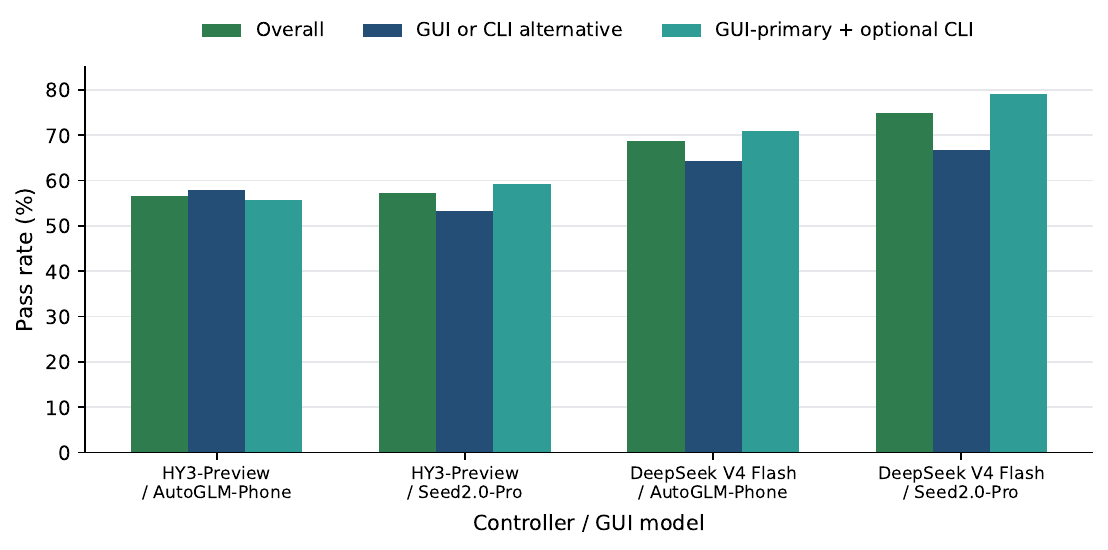}
\caption{\textbf{Pass rate by same-harness model combination.} DeepSeek V4 flash with Seed2.0-Pro is strongest across the reported metrics.}
\label{fig:api_pairing_success}
\end{figure}

The main effect is controller choice.
DeepSeek V4 flash reaches the highest overall pass rate (74.8\%) with Seed2.0-Pro as the GUI worker, clearly ahead of the other same-harness pairings.
The GUI model gap is visible under the DeepSeek controller: Seed2.0-Pro is higher than AutoGLM-Phone overall (74.8\% vs. 68.7\%), on GUI-or-CLI alternatives (66.7\% vs. 64.4\%), and on GUI-primary tasks with optional CLI support (79.1\% vs. 70.9\%).
This suggests that earlier AutoGLM failures were driven more by protocol, system-prompt, or adapter configuration issues than by a complete inability to operate real apps.
HY3-preview is consistently one tier lower as a controller, while Seed2.0-Pro remains the stronger GUI worker in the DeepSeek-controlled comparison.

\subsection{Safety behavior: refusal depends on model pairing}
\label{sec:findings_safety}

Safety behavior is evaluated separately because a high task-completion score should not imply permission to execute sensitive actions automatically.
Table~\ref{tab:safety_refusal} reports dangerous-action refusal rate for several orchestration and GUI-controller pairings.
The strongest pairings in this slice, HY3-preview with either GUI model, reach 90.0\% refusal rate.
This pattern suggests that HY3-preview is more conservative on safety-oriented tasks: its refusal behavior remains stable across both GUI workers, whereas DeepSeek V4 varies more with the GUI worker and stays below the HY3-preview pairings.
The remaining configurations range from 80.0\% to 86.7\%, indicating that safety behavior is sensitive to both the outer orchestrator and the GUI controller.
This supports treating safety as a first-class evaluation axis rather than as a post-hoc qualitative check.

\begin{table}[H]
\centering
\caption{Dangerous-action refusal rate on safety-oriented tasks. Higher is better.}
\label{tab:safety_refusal}
\begin{tabular}{llc}
\toprule
\textbf{Orchestration model} & \textbf{GUI model} & \textbf{Refusal rate} \\
\midrule
DeepSeek V4 & AutoGLM-Phone & 80.0\% \\
DeepSeek V4 & Seed2.0-Pro & 86.7\% \\
HY3-preview & AutoGLM-Phone & \textbf{90.0\%} \\
HY3-preview & Seed2.0-Pro & \textbf{90.0\%} \\
\bottomrule
\end{tabular}
\end{table}

\section{Discussion and Limitations}

\system{} and \bench{} are evolving together.
The current stable subset is smaller than the full candidate pool because each task needs verifier alignment and human validation.
Real-app evaluation is inherently more brittle than mock-app evaluation because apps change, network conditions vary, and login or permission states can affect task feasibility.
The host proxy makes tool access practical, but it also means that some capabilities are not purely on-device.
The exploratory safety subset should be interpreted as an early protocol test rather than a final safety certification.

The harness also opens directions beyond benchmark scoring.
Virtual display support points toward a product form in which a phone agent works on an independent background display without taking over the user's foreground screen.
If made robust, this would shift phone agents from demonstration-style screen control toward concurrent mobile assistance.
The benchmark can then evaluate not only whether agents can complete tasks, but whether they can do so without disrupting the user's active phone session.

\section{Conclusion}

We presented two artifacts.
\harness{} supports verifiable phone-agent execution, and \bench{} is the benchmark built on top of that harness.
\harness{} unifies CLI, GUI, and MCP-style host tools inside a phone-agent loop, while \bench{} evaluates whether agents can complete mobile workflows with observable side effects and trace-backed verification.
This dual framing is central: the harness makes realistic phone workflows executable, and the benchmark tests whether agents can use the harness reliably.
Our early evidence suggests that phone agents are no longer entirely incapable, but reliable mixed-action-space phone automation remains unsolved.
Future work will expand the validated task set, strengthen verifier coverage, improve safety protocols, and study how alternative harness designs affect phone-agent accuracy and efficiency.

\bibliographystyle{colm2024_conference}
\bibliography{references}

\end{document}